\documentclass[10pt,twocolumn,letterpaper]{article}

\usepackage{iccv}
\usepackage{times}
\usepackage{epsfig}
\usepackage{graphicx}
\usepackage{amsmath}
\usepackage{amssymb}

\usepackage[accsupp]{axessibility}  

\usepackage{array}
\usepackage{booktabs}
\usepackage{bbding} 
\usepackage{bm}
\usepackage{enumerate}
\usepackage{multirow}
\usepackage{multicol}
\usepackage{makecell}
\usepackage{colortbl}
\usepackage{color}

\usepackage[pagebackref=true,breaklinks=true,letterpaper=true,colorlinks,bookmarks=false]{hyperref}

\usepackage[capitalize]{cleveref}
\crefname{section}{Sec.}{Secs.}
\Crefname{section}{Section}{Sections}
\Crefname{table}{Table}{Tables}
\crefname{table}{Tab.}{Tabs.}

\newcommand{\subtitle}[1]{{\noindent}{\textbf{#1}}}

\usepackage{xcolor}

\definecolor{myPurple}{rgb}{0.4, .0, .8}
\definecolor{myGreen}{rgb}{0, .8, .3}
\definecolor{myRed}{rgb}{0.8, .2, .2}
\definecolor{myOrange}{rgb}{0.7, 0.45, 0.2}
\definecolor{myBlue}{rgb}{.0, .0, 1.0}
\definecolor{myBlue2}{rgb}{.0, .0, 0.5}
\definecolor{myBlack}{rgb}{.0, .0, 0.0}

\iccvfinalcopy 

\def\iccvPaperID{6105} 

\ificcvfinal\fi

\begin{document}


\title{Minimum Latency Deep Online Video Stabilization}
\author{
    Zhuofan Zhang\textsuperscript{\rm1\footnotemark[1]} \qquad Zhen Liu\textsuperscript{\rm2\thanks{Equal contribution}} \qquad Ping Tan\textsuperscript{\rm3} \qquad Bing Zeng\textsuperscript{\rm1} \qquad Shuaicheng Liu\textsuperscript{\rm1,\rm2\thanks{Corresponding author}}\\
    \textsuperscript{\rm1}University of Electronic Science and Technology of China \ \textsuperscript{\rm2}Megvii Technology \\
    \textsuperscript{\rm3}The Hong Kong University of Science and Technology \\
    \tt\small\{zhangzf98@std., eezeng@, liushuaicheng@\}uestc.edu.cn, \\ \tt\small liuzhen03@megvii.com, \tt\small pingtan@ust.hk
}

\maketitle
\ificcvfinal\thispagestyle{empty}\fi

\begin{abstract}
   We present a novel camera path optimization framework for the task of online video stabilization. Typically, a stabilization pipeline consists of three steps: motion estimating, path smoothing, and novel view rendering. Most previous methods concentrate on motion estimation, proposing various global or local motion models. In contrast, path optimization receives relatively less attention, especially in the important online setting, where no future frames are available. In this work, we adopt recent off-the-shelf high-quality deep motion models for motion estimation to recover the camera trajectory and focus on the latter two steps. Our network takes a short 2D camera path in a sliding window as input and outputs the stabilizing warp field of the last frame in the window, which warps the coming frame to its stabilized position. A hybrid loss is well-defined to constrain the spatial and temporal consistency. In addition, we build a motion dataset that contains stable and unstable motion pairs for the training. Extensive experiments demonstrate that our approach significantly outperforms state-of-the-art online methods both qualitatively and quantitatively and achieves comparable performance to offline methods. Our code and dataset are available at \url{https://github.com/liuzhen03/NNDVS}.
\end{abstract}


\section{Introduction}
Video stabilization methods aim at removing unwanted shaky motions of a video caused by unsteady moving capture~\cite{guilluy2021video}. Traditional methods often take three main steps: 1) camera motion estimating for trajectory recovery; 2) camera path smoothing; and 3) steady frame synthesis. According to the adopted motion model in the first step, these methods can be broadly classified as 2D or 3D. 2D methods adopt planar motion models such as affine~\cite{grundmann2011auto}, or homography~\cite{matsushita2006full}, or mesh~\cite{liu2013bundled, liu2016meshflow, wang2018deep}, or flow~\cite{liu2014steadyflow,liu2017codingflow,yu2020learning}, while 3D methods resort to 3D depth~\cite{liu2012video,lee20213d}, or reconstructed points~\cite{liu2009content}, or epipolar geometry constraints~\cite{goldstein2012video}.

\begin{figure}[t]
   \centering
   \includegraphics[width=1.0\linewidth]{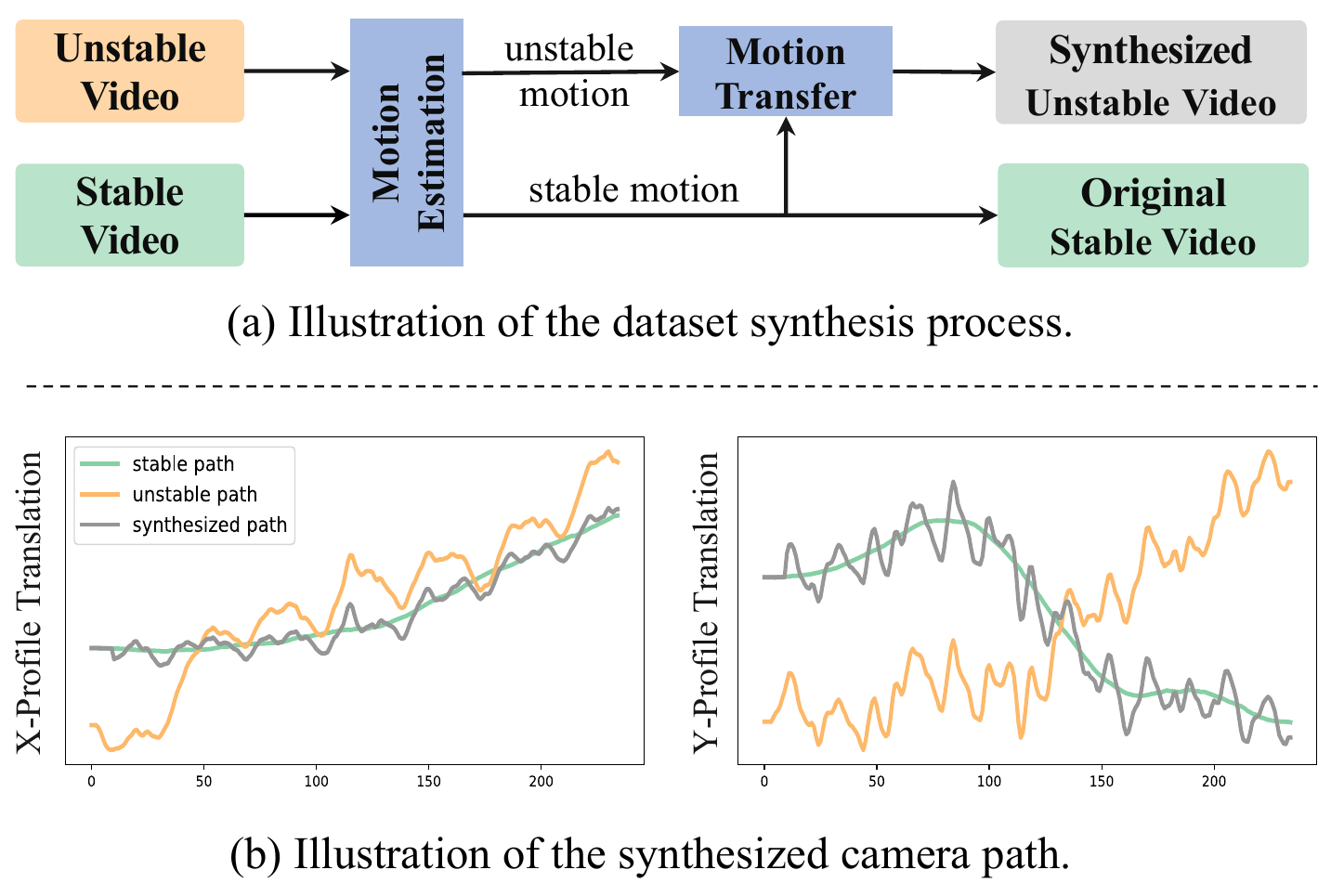}   
   \caption{Illustration of the dataset synthesis process (a) and the synthesized camera path (b). We transfer an unstable path to a stable path, creating a synthesized path, which shares high frequencies of the unstable path with low frequencies of the stable path.}\label{fig:teaser}
 \end{figure}
   
In comparison, deep learning-based methods learn stabilization models from stable and unstable video pairs~\cite{wang2018deep, yu2020learning, xu2018deep} without explicit steps of motion estimation and smoothing. However, the results of deep methods are often visually inferior to those of traditional ones. A potential reason is that these methods try to learn the three steps all in a unified framework, each of which has different characteristics but is learned in the network blindly.      

On the other hand, deep affine~\cite{jin2021deep}, deep homography~\cite{Ye_2021_ICCV,hong2022unsupervised}, and deep mesh models~\cite{ye2019deepmeshflow,liu2022content,liu2022unsupervised,Nie9605632} have demonstrated high-quality image registration, even under adverse cases. They are robust to scenes with large depth variation, large dynamic objects, and poor textures, fitting perfectly for the camera motion estimation. In this work, we argue that it is not necessary to enclose all three steps of the stabilization into the learning pipeline, but let the motion estimation to the recent deep motion models~\cite{ye2019deepmeshflow,liu2022content}. In this way, our network only focuses on learning to stabilize the shaky camera motion, which makes the learning process much more efficient and effective.

Camera path smoothing can be performed offline or online, where the former optimizes the path globally while the latter smooths the path on-the-fly. In other words, offline approaches have access to all past and future frames during optimization, as it often stabilizes a video after it is captured. In contrast, online methods aim to stabilize a video during the capture. Note that, the concept of real-time differs from that of online, as an online method must be in real-time whereas an offline method can run at real-time speed. The main distinction is the availability of future frames. 

Most existing methods are offline~\cite{liu2013bundled,grundmann2011auto,grundmann2012calibration,wang2013spatially,liu2014steadyflow}. However, the online setting is critical as many applications desire instant visual feedback based on live video streams~\cite{liu2016meshflow,wang2018deep}. Normally, with limited or even no access to future frames, online methods cannot achieve equal stability as offline methods, particularly in the suppression of the low-frequency camera shake, which frequently necessitates a long camera path. One may argue that online methods can have all the past frames for processing. However, future motions are important, if not more important than the past. Because, for one thing, fast camera motions can happen at any time suddenly, e.g., quick rotation or zooming. The reaction time is short. Inappropriate processing may either decrease the stability or create artifacts such as excessive cropping. For another, all the past frames have already been shown to the audience, meaning that they are fixed, and thus cannot be modified. We can only adjust the unstabled future frames to the frozen past ones.

To this end, we propose a deep online camera path smooth network that takes a pre-estimated short 2D camera trajectory in a sliding window as input and outputs the motion compensation warp field of the last frame in the window for the stabilized view, which is the setting of online stabilization with the minimum latency. A video can be stabilized on the fly by applying our model repeatedly for every incoming frame. 
To deal with depth changes and moving objects, we employ the mesh-based motion model~\cite{ye2019deepmeshflow} which demonstrates excellent robustness in various scenes. 
Instead of directly inputting the mesh structure or mesh cells as local homographies to the network, we convert the estimated motion to a dense flow field, which can be fed into the convolutional neural network naturally. We show that it is a more convenient representation compared to other alternatives, e.g., vertex sparse motion vectors or homography matrix, which cannot be easily adapted as input to CNNs. Based on that, we further propose a hybrid loss, which consists of the motion-consistency loss, the shape-consistency loss, and the scale-preserving loss, to maintain the spatial coherence and temporal continuity of the stabilized video.

Besides, we create a motion dataset, \textbf{MotionStab}, to train our network. In particular, we capture 110 stable videos with a cell phone mounted on a hand-held physical stabilizer. Then, we create a shaky video by transforming each frame of the stable video according to the motion of another irrelevant shaky video. In this way, the stable and unstable motion pairs can be constructed. Fig.~\ref{fig:teaser} shows our idea. Note that, our dataset is different from DeepStab~\cite{wang2018deep}, which is captured by a stable and a shaky camera simultaneously. We directly transfer the motion of a shaky video to a stable video. The contents of the two videos are not important but the motion does. That means DeepStab~\cite{wang2018deep} offers frame pairs, while our MotionStab offers motion pairs. We feed the motions to the network instead of RGB frames.

The benefits are three folds: 1) image contents are diverse, while motion mappings are much easier to learn; 2) motion as mesh is lightweight, so does the network; 3) the network can be more focused by excluding the task of the motion estimation. As a result, we show that this motion learning pipeline is effective enough, such that it can work with straightforward network architectures, e.g., UNet, without help from more advanced
designs or sophisticated modules. Complex motion types can be learned by the network successfully, as long as they exist in the training data. As such, the network can be kept simple yet effective. 

In summary, the main contributions of this paper are concluded as follows:
\begin{itemize}
   \item We propose a novel deep camera path optimization framework for online video stabilization. 
   
   \item We propose a hybrid loss to enable robust supervision for maintaining the spatial and temporal coherence of the stabilized video.
   
   \item We build a comprehensive MotionStab dataset that covers various motions for camera path optimization.
   
   \item We conduct extensive experiments to demonstrate the effectiveness of the proposed approach against existing state-of-the-art online and offline methods.
\end{itemize}

\section{Related Works}

\subsection{Traditional Methods}
Early 2D approaches track image features for several frames and then smooth these feature trajectories to stabilize a video~\cite{lee2009video,wang2013spatially}. However, long feature tracks~\cite{shi1994good} are hard to be obtained when large camera motions exist. Later, motion models, such as affine~\cite{grundmann2011auto} or homography~\cite{matsushita2006full,gleicher2008re}, are calculated between neighboring frames, relaxing the requirement of long feature tracks into feature matches~\cite{rublee2011orb} between adjacent frames. Then, these motion models are accumulated to generate a 2D camera trajectory, which is shown to be a good replacement against track-based approaches, owning largely improved robustness. With respect to the motion representation, homography mixture~\cite{grundmann2012calibration}, mesh~\cite{liu2013bundled, liu2016meshflow,wang2013spatially},  and optical flow~\cite{liu2014steadyflow,liu2017codingflow} based motion models are proposed to deal with scenes that contain large depth variations. 
Besides, some approaches focus on special stabilization tasks, such as Selfie~\cite{yu2021real}, 360~\cite{kopf2016360}, and hyperlapse videos~\cite{joshi2015real}. 

On the other hand, 3D-based methods require 3D camera motions or scene structure for stabilization. The 3D structure can either be calculated from the video by structure-from-motion (SfM)~\cite{liu2009content} or acquired from additional hardware, such as a depth camera~\cite{liu2012video}, a gyroscope sensor~\cite{karpenko2011digital}, or a light field camera~\cite{smith2009light}. However, full 3D stabilization is fragile and computationally expensive~\cite{liu2009content}. Partial 3D methods proposed 3D constraints, such as subspace projection~\cite{liu2011subspace} and epipolar geometry~\cite{goldstein2012video}, to alleviate the requirement of full 3D reconstruction~\cite{hartley2003multiple}. Normally, 3D methods can better handle the scene parallax compared with 2D approaches due to the physical correctness, as long as the 3D information can be available. 

\begin{figure*}[t]
   \centering
   \includegraphics[width=0.95\linewidth]{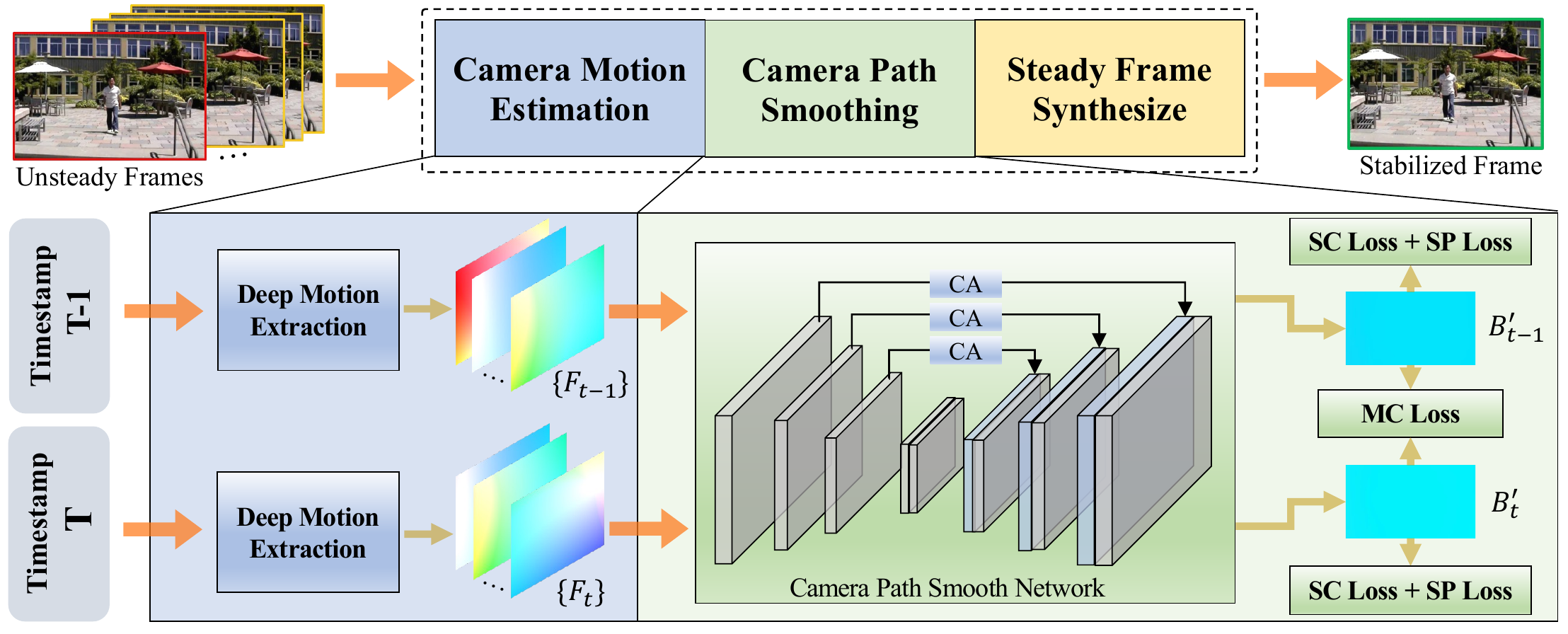}   
   \caption{The overall pipeline of the proposed deep online camera path optimization framework. (a) We first employ a deep motion extraction model to estimate the unsteady camera motion. (b) Then, the estimated motion is fed into the camera path smooth network, yielding the smoothed warp field. (c) Finally, the target steady frame is synthesized by the predicted smooth warp field.}\label{fig:pipeline}
 \end{figure*}

\subsection{Deep Learning Methods}
Deep methods take the video frames as input and directly output the stabilized frames, which are often trained with stable and unstable frame pairs acquired by special hardware, e.g., DeepStab dataset~\cite{wang2018deep}. Xu~\emph{et al.} used the adversarial network to generate a target image to guide the warping~\cite{xu2018deep}. Yu~\emph{et al.} learned the flow fields from initial optical flow estimation for the accurate per-pixel motion compensation~\cite{yu2020learning}. Deep methods suffer from the generalization problem, given that DeepStab only contains 60 videos. Yu~\emph{et al.} used the CNN as the optimizer instead of learning from data to overfit each input example~\cite{yu2019robust}. 

Recently, deep motion models have achieved good results for motion estimation, such as deep homography~\cite{Ye_2021_ICCV, hong2022unsupervised} and deep meshflow~\cite{ye2019deepmeshflow}, which is more robust in adverse cases compared to traditional solutions, such as scenes of low texture and low light. In this work, we adopt the deep meshflow~\cite{ye2019deepmeshflow} for our camera motion estimator and design a network concentrating on camera path smoothing.

\subsection{Online Approaches}
Most of the video stabilization approaches are offline, where a video is processed after it has been captured. Online methods stabilize a video during the capture~\cite{yang2006online,wang2018deep,liu2016meshflow}.  
One may argue that global path optimization can be easily modified to online methods by applying a sliding window scheme. However, the stability would inevitably decrease, let alone the existence of challenging camera motions, such as quick rotation and zooming, which may introduce artifacts such as wobble and excessive cropping. Previously, these difficult motions can only be solved satisfactorily by global camera path smoothing. In this work, we propose to solve these problems with our proposed deep online camera path optimization framework.

\section{Methodology}

\subsection{Problem Formulation}
Our method is built upon convolutional neural networks. As illustrated in Fig.~\ref{fig:camera_motion} (a), given an incoming unsteady frame $I_{t}$ at timestamp $t$, our deep online video stabilization aims to predict the corresponding steady frame ${I}_{t}^{'}$ with no future frames. Here, we use `$'$' to represent predicted quantities. Note that, previous learning-based methods learn the mapping $f(\cdot)$ from the input shaky RGB frames and predict a warp field ${B}_{t}^{'}$ of frame $I_{t}$, which can be formulated as 
\begin{align}
   \{B_{t}^{'}\} = f \left(\{I_{t}\} \right),
\end{align}
where `$\{\cdot\}$' represents a set of elements. However, our experiments show that directly using complex RGB video frames as input and mixing motion estimation and path smoothing in a network frequently causes estimation errors, resulting in wobbling and distortion artifacts. 

In this work, we propose our approach from a novel perspective, where we separate the motion estimation step from the network alone and leave the network focus on camera path smoothing. Specifically, we consider using a fixed window of $r$ past frames $\{I_{t}\}_r=\left\langle{}I_{t-r}, ..., I_{t-1}, I_{t}\right\rangle$ of minimum latency (i.e. without using future frames) to stabilize the incoming target frame $I_t$. The camera motion $\{F_{t}\}$ is estimated by an off-the-shelf deep motion estimation model. Our camera path smooth network uses only the estimated motion as input to predict the corresponding warp field $B_{t}^{'}$, i.e.,
\begin{align}\label{eq:formulation}
   \{B_{t}^{'}\} = \Phi (\{F_{t}\}; \theta),
\end{align}
where $\Phi(\cdot)$ denotes the camera path smoothing network and $\theta$ is the network parameters to be optimized. We empirically set $r$ to 15 in our experiments.

\subsection{Deep Online Camera Path Optimization}
As shown in Fig.~\ref{fig:pipeline}, the overall pipeline of the proposed deep online camera path optimization framework mainly consists of three stages. Firstly, we adopt a deep motion extraction model for robust camera motion estimation. Secondly, the camera path smooth network takes the unsteady camera motion as input and predicts the smoothed warp field. Finally, the original shaky frame is transformed by the predicted warp field to synthesize the target steady frame.

\begin{figure}[t]
   \centering
   \includegraphics[width=1.0\linewidth]{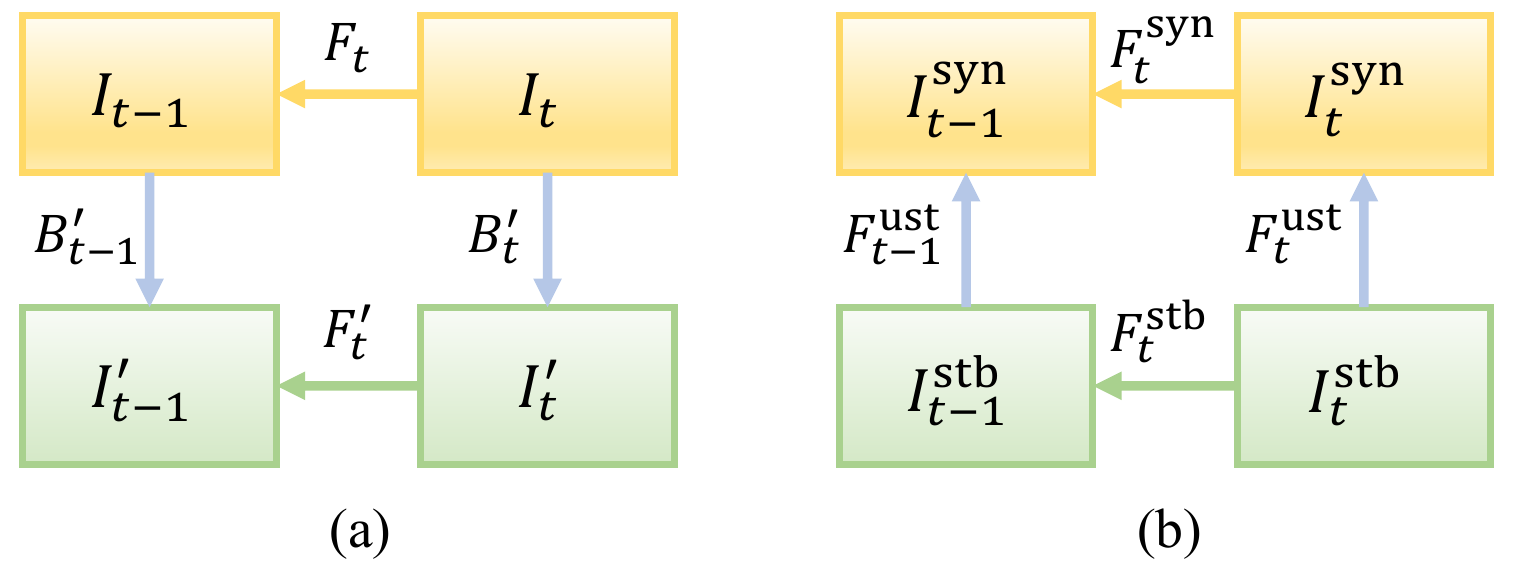}   
   \caption{Relationships between the unstable/synthesized video and the stable video. (a) depicts the basic relationships of the unstable video $\{I_{t}\}$ and stabilized video $\{I^{'}_{t}\}$, and (b) illustrates the main idea of our dataset synthesis strategy that transfers collected unstable motions $\{F^{\text{ust}}_{t}\}$ to stable video frames $\{I^{\text{stb}}_{t}\}$, generating synthesized unstable video frames $\{I^{\text{syn}}_{t}\}$.}\label{fig:camera_motion}
 \end{figure}

~\\
\subtitle{Camera Motion Estimation} The first step is to estimate the camera motion. We employ an optimized deep meshflow model~\cite{ye2019deepmeshflow} to extract the inter-frame motion. Specifically, given the incoming frame $I_t$ and past adjacent frame $I_{t-1}$ with size of $640\times 360$, we estimate $F_t$ by model~\cite{ye2019deepmeshflow}  (Fig.~\ref{fig:camera_motion} (a), upper yellow part). Repeat this process, we accumulate a set of inter-frame motions, $\{F_{t}\}_{r-1} = \left\langle{}F_{t-r+1}, ..., F_{t-1}, F_{t}\right\rangle$. In particular, we keep the most recent $r-1$ motions in our buffer, where historical motions that are beyond range $r-1$ are simply dropped. The deep meshflow motion estimation is defined as: 
\begin{align}
\label{eq:f_t}
   F_{t} = \downarrow_{s}\mathcal{M}_{F}(I_{t}, I_{t-1}),
\end{align}
where $\mathcal{M}_{F}(\cdot)$ is the pre-trained deep meshflow model. `$\downarrow_{s}$' indicates that we collect the sparse motions at the mesh vertexes, and shrink them pixel by pixel to form a downsampled dense motion field, for the purpose of feeding it to the CNNs. The scale factor $s$ is set to 8, and the size of the dense motion field $F_{t}$ is $80\times 45$.

~\\
\subtitle{Camera Path Smooth Network} Given the estimated camera motion $\{F_{t}\}_{r-1}$ as input, the camera path smooth network predicts the target smoothed warp field $B_{t}^{'}$ as Eq.~\ref{eq:formulation}. Note that, the meshflow model is spatially-variant, and so does the $F_t$. Therefore, we provide spatial and temporal optimizations. The camera path smooth network is designed as an encoder-decoder architecture and we implement it as a 4-stage UNet with channels of 64, 128, 256, and 512. We further employ the channel attention mechanism (CA)~\cite{hu2018squeeze} to adaptively learn the weights of different historical motions. 

~\\
\subtitle{Steady Frame Rendering} After obtaining the warp field $B_{t}^{'}$, we inverse the process of shrinking, where we put the motions back to the sparse vertex locations to create a mesh, based on which we warp the shaky frame $I_{t}$ to its stabilized position $I'_{t}$:
\begin{align}
I'_{t} = \mathcal{W}(I_{t}, B_{t}^{'}),
\end{align}
where $\mathcal{W}(\cdot)$ denotes the backward warping function. The first parameter is an image and the second is a warp field.

\subsection{Training Dataset}

Existing training datasets are either captured simultaneously by two cameras~\cite{wang2018deep} or synthesized by simulated injected noise~\cite{huang2019stablenet,qu2013shaking, ito2019dataset}. The former would naturally suffer from parallax problems, while the latter makes the synthesized video paths unrealistic. To address this issue, we offer a novel synthetic method that can build realistic unstable/stable motion pairs from existing shaky videos.

Our main idea is to transfer the motion of an \textbf{unstable} shaky video $\boldsymbol{V}_{\text{ust}}$ to a \textbf{stable} video $\boldsymbol{V}_{\text{stb}}$, yielding another \textbf{synthesized} shaky video $\boldsymbol{V}_{\text{syn}}$ with motions from $\boldsymbol{V}_{\text{ust}}$ but image contents as $\boldsymbol{V}_{\text{stb}}$. The videos $\boldsymbol{V}_{\text{ust}}$ and $\boldsymbol{V}_{\text{stb}}$ are irrelevant. Let's denote $\{I_t^{\text{ust}}\}, \{I_t^{\text{
stb}}\}, \{I_t^{\text{syn}}\}$ as video frames and $\{F_t^{\text{ust}}\}, \{F_t^{\text{stb}}\}, \{F_t^{\text{syn}}\}$ as motion between frames of video $\boldsymbol{V}_{\text{ust}}, \boldsymbol{V}_{\text{stb}}, \boldsymbol{V}_{\text{syn}}$, accordingly. 

To achieve this, we first estimate the motions $\{F_t^{\text{ust}}\}$, $\{F_t^{\text{stb}}\}$ between adjacent frames of $\boldsymbol{V}_{\text{ust}}$ and $\boldsymbol{V}_{\text{stb}}$ by deep meshflow~\cite{ye2019deepmeshflow}. Next, we warp the frame $\{I_t^{\text{stb}}\}$ by motion $\{F_t^{\text{ust}}\}$ to produce $\{I_t^{\text{syn}}\}$, for each timestamp $t$:
\begin{align}\label{eq:data_tran}
   I_t^{\text{syn}} = \mathcal{W}(I_t^{\text{stb}},F_t^{\text{ust}}).
\end{align}

Applying Eq.~\ref{eq:data_tran} to every frame of $\boldsymbol{V}_{\text{stb}}$ creates the desired $\boldsymbol{V}_{\text{syn}}$. The $\boldsymbol{V}_{\text{stb}}$ and $\boldsymbol{V}_{\text{syn}}$ become a pair of stable/unstable videos, where the following motion equality holds (Please refer to Fig.~\ref{fig:camera_motion} (b) for the relation), for each timestamp $t$:
\begin{align}\label{eq:circle}
F_{t}^{\text{ust}} + F_{t}^{\text{syn}} = F_{t}^{\text{stb}} + F_{t-1}^{\text{ust}},
\end{align}
where $F_t^{\text{ust}}$, $F_t^{\text{stb}}$ have been calculated which are known. Therefore, $F_{t}^{\text{syn}}$ can be calculated as:
\begin{align}\label{eq:circle2}
F_{t}^{\text{syn}} = F_{t}^{\text{stb}} + F_{t-1}^{\text{ust}} - F_{t}^{\text{ust}}.
\end{align}
During the training, our network takes a subset of motions $\{F_t^{\text{syn}}\}_{r-1} = \left\langle{}F_{t-r+1}^{\text{syn}}, ...,  F_{t-1}^{\text{syn}}, F_{t}^{\text{syn}}\right\rangle$ as input and output a motion field that is as close as the ground-truth label `$-F_t^{\text{ust}}$', which is the inverse motion of `$F_t^{\text{ust}}$' as shown in Fig.~\ref{fig:camera_motion} (b). On the other hand, the generated videos $\boldsymbol{V}_{\text{stb}}$ and $\boldsymbol{V}_{\text{syn}}$ are not important, as all we need are motions that can be calculated and derived by relations.

To further establish the training dataset, we collected 110 stable videos of five categories (including regular, rotation, zooming, parallax, and crowd) using a cell phone mounted on a hand-held physical stabilizer. Next, we collected over 150 unstable videos from the website. Subsequently, we performed data synthesis on the collected stable and unstable videos by employing the aforementioned method, constructing the \textbf{MotionStab} dataset, which contains paired unstable/stable camera motion of various complex scenarios. As illustrated in Fig.~\ref{fig:teaser} (b), the camera path of the synthesized video pairs is well-registered and realistic as the original unstable one.

\subsection{Training Loss}
To train the camera path smooth network, we define several losses in our framework: the motion-consistency loss for maintaining motion continuity between consecutive frames, the shape-consistency loss for preserving the geometry of the predicted warp field, and the scale-preserving loss for maintaining the scale of the predicted warp field. 

~\\
\subtitle{Motion-consistency Loss} To keep the continuity of the predicted consecutive video frames, the inter-frame motion of the stabilized video $F_{t}^{'}$ should be as similar as the ground truth one $\hat{F}_{t}$, i.e.,
\begin{align}\label{eq:mc_loss_1}
   \mathcal{L}_{\text{MC}} = \left\| F_{t}^{'} - \hat{F}_{t} \right\|_{1}.
\end{align}
According to the relationship of Fig.~\ref{fig:camera_motion} (a), Eq.~\ref{eq:mc_loss_1} can be written as
\begin{align}\label{eq:mc_loss_2}
   \mathcal{L}_{\text{MC}} &= \left\| (F_{t} + B_{t-1}^{'} - B_{t}^{'}) - (F_{t} + \hat{B}_{t-1} - \hat{B}_{t})  \right\|_{1} \\
                           &= \left\|  B_{t-1}^{'} - B_{t}^{'} - \hat{B}_{t-1} + \hat{B}_{t} \right\|_{1},
\end{align}
where $B_{t}^{'}$ denotes network output of the predicted warp field. That is to say, the motion continuity can be restrained by the ground truth warp field $\hat{B}_{t}$ and $\hat{B}_{t-1}$.

\begin{figure}[t]
   \centering
   \includegraphics[width=1.0\linewidth]{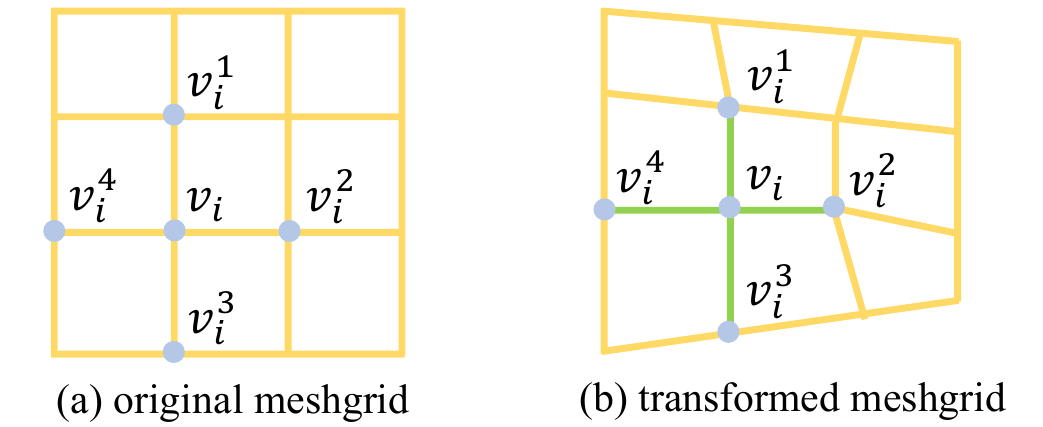}   
   \caption{Illustration of the shape-consistency loss and the scale-preserving loss. The shape-consistency loss is employed to preserve the geometry shape of a meshgrid and the scale-preserving loss is used to maintain the scale.}\label{fig:loss}
 \end{figure}

\begin{table*}[t]
   \centering
   \caption{Quantitative comparisons on the NUS dataset~\cite{liu2013bundled}. We use the Cropping ratio (C), the Distortion value (D), and the Stability score (S) as evaluation metrics. All of these metrics are in the range of 0 to 1, and the higher the better. The `$\ast$' denotes the results of Bundled~\cite{liu2013bundled} are computed directly from the NUS dataset. The bests are marked in \textcolor{red}{red} and the 
  second bests are in \textcolor{blue}{blue}.}
   \label{tab:quantitative_results_nus}
   \resizebox*{0.95\linewidth}{!}{
      \begin{tabular}{
         >{\arraybackslash}p{1.2cm}| 
				>{\centering\arraybackslash}p{2.6cm}| 
				   >{\centering\arraybackslash}p{1.0cm} 
				>{\centering\arraybackslash}p{1.0cm} 
				>{\centering\arraybackslash}p{1.0cm}| 
    				>{\centering\arraybackslash}p{1.0cm} 
				>{\centering\arraybackslash}p{1.0cm} 
				>{\centering\arraybackslash}p{1.0cm}| 
    				>{\centering\arraybackslash}p{1.0cm} 
				>{\centering\arraybackslash}p{1.0cm} 
				>{\centering\arraybackslash}p{1.0cm} 
      }
      \toprule
      \hline
      \multirow{2}{*}{} & \multirow{2}{*}{Methods} & \multicolumn{3}{c|}{Regular} & \multicolumn{3}{c|}{Rotation} & \multicolumn{3}{c}{Zooming} \\
      &  & C & D & S & C & D & S & C & D & S \\
      \hline
         \multirow{3}{*}{Offline}  & $\ast$ Bundled~\cite{liu2013bundled} & 0.6658 & 0.9409 & \textcolor{red}{0.9048} & 0.6477 & 0.8692 & 0.9194 & 0.5773 & 0.9073 & \textcolor{red}{0.9236} \\
          & Robust L1~\cite{grundmann2011auto} & 0.7157 & 0.9252 & 0.8454 & 0.7206 & 0.8231 & 0.8617 & 0.7107 & 0.8329 & 0.7730 \\
          & DIFRINT~\cite{choi2020deep} & \textcolor{red}{0.9854} & 0.9555 & 0.8177 & \textcolor{red}{0.9413} & 0.8813 & 0.8701 & \textcolor{red}{0.9528} & 0.8531 & 0.8763 \\
          & Yu~\emph{et al.}~\cite{yu2020learning} & \textcolor{blue}{0.9486} & 0.9752 & 0.8427 & 0.7661 & 0.7616 & \textcolor{blue}{0.9317} & 0.8971 & 0.8630 & 0.9107 \\
          & PWStableNet~\cite{zhao2020pwstablenet} & 0.9319 & \textcolor{blue}{0.9852} & 0.8162 & 0.8801 & \textcolor{blue}{0.9707} & 0.9251 & \textcolor{blue}{0.9303} & \textcolor{red}{0.9792} & 0.8906 \\
          & DUT~\cite{xu2022dut} & 0.9485 & 0.9600 & \textcolor{blue}{0.8717} & 0.7609 & 0.6940 & 0.9203 & 0.8986 & 0.8606 & \textcolor{blue}{0.9221} \\
      \hline
         \multirow{3}{*}{Online}  & Meshflow~\cite{liu2016meshflow} & 0.8081 & 0.9168 & 0.8386 & 0.7578 & 0.7679 & 0.9092 & 0.7773 & 0.8715 & 0.8839 \\
          & StabNet~\cite{wang2018deep} & 0.7491 & 0.8393 & 0.8406 & 0.7417 & 0.7205 & 0.8384 & 0.7376 & 0.7807 & 0.8697 \\
          & Ours & 0.9433 & \textcolor{red}{0.9917} & 0.8030 & \textcolor{blue}{0.9021} & \textcolor{red}{0.9847} & \textcolor{red}{0.9341} & 0.9185 & \textcolor{blue}{0.9736} & 0.8982 \\
      \hline
      \hline
      \multirow{2}{*}{} & \multirow{2}{*}{Methods} & \multicolumn{3}{c|}{Crowd} & \multicolumn{3}{c|}{Parallax} & \multicolumn{3}{c}{Avg.} \\
      &  & C & D & S & C & D & S & C & D & S \\
      \hline
         \multirow{3}{*}{Offline}  & $\ast$ Bundled~\cite{liu2013bundled} & 0.6685 & 0.8910 & \textcolor{blue}{0.8744} & 0.7158 & 0.8965 & \textcolor{blue}{0.8964} & 0.6550 & 0.9010 & \textcolor{blue}{0.9037} \\
          & Robust L1~\cite{grundmann2011auto} & 0.7208 & 0.9077 & 0.8504 & 0.7132 & 0.8054 & 0.8778 & 0.7162 & 0.8589 & 0.8417 \\
          & DIFRINT~\cite{choi2020deep} & \textcolor{red}{0.9662} & 0.8713 & 0.8456 & \textcolor{red}{0.9746} & 0.8823 & 0.8600 & \textcolor{red}{0.9641} & 0.8887 & 0.8539 \\
          & Yu~\emph{et al.}~\cite{yu2020learning} & 0.9081 & 0.9191 & 0.8660 & 0.9209 & 0.9219 & 0.8896 & 0.8882 & 0.8872 & 0.8881 \\
          & PWStableNet~\cite{zhao2020pwstablenet} & 0.9181 & \textcolor{blue}{0.9790} & 0.8162 & 0.9244 & \textcolor{blue}{0.9802} & 0.8549 & 0.9170 & \textcolor{blue}{0.9789} & 0.8606 \\
          & DUT~\cite{xu2022dut} & 0.8986 & 0.8617 & \textcolor{red}{0.9042} & 0.9254 & 0.8774 & \textcolor{red}{0.9127} & 0.8860 & 0.8507 & \textcolor{red}{0.9062} \\
      \hline
         \multirow{3}{*}{Online}  & Meshflow~\cite{liu2016meshflow} & 0.7888 & 0.8246 & 0.8280 & 0.8121 & 0.8429 & 0.8488 & 0.7882 & 0.8762 & 0.8617 \\
          & StabNet~\cite{wang2018deep} & 0.7247 & 0.7725 & 0.7912 & 0.6967 & 0.7660 & 0.8289 & 0.7280 & 0.7758 & 0.8338 \\
          & Ours & \textcolor{blue}{0.9247} & \textcolor{red}{0.9816} & 0.8408 & \textcolor{blue}{0.9326} & \textcolor{red}{0.9827} & 0.8762 & \textcolor{blue}{0.9242} & \textcolor{red}{0.9829} & 0.8705 \\
      \hline
      \bottomrule
      \end{tabular}}
\vspace{-0.5em}
\end{table*}

~\\
\subtitle{Shape-consistency Loss}
The shape-consistency loss is effective to prevent the distortion problem of the predicted warp field. Following Wang \emph{et al.}~\cite{wang2018deep}, we implement the SC loss as an intra-grid loss term and an inter-grid loss term:
\begin{align}\label{eq:sc_loss}
   \mathcal{L}_{\text{SC}} = \mathcal{L}_{\text{intra}} + \mathcal{L}_{\text{inter}}.
\end{align}
Note that each pixel of our dense warp field is actually a vertex of the mesh grid. As depicted in Fig.~\ref{fig:loss}, the intra-grid loss term is defined as: 
\begin{align}\label{eq:intra_grid_loss}
   \mathcal{L}_{\text{intra}} =  \frac{1}{N} \sum_{v_i}\left\|  (v_{i}^{1} - v_{i})\cdot (v_{i}^{2} - v_{i}) \right\|_{1},
\end{align}
where $i$ denotes the $i$-th vertice and $N$ is the total number of vertices. Besides, the inter-grid loss term is used to maintain the geometric consistency of adjacent grids:
\begin{align}\label{eq:inter_grid_loss}
   \mathcal{L}_{\text{inter}} = \frac{1}{N} \sum_{v_i}\left\| (v_{i}^{1} - v_{i}) - (v_{i} - v_{i}^{3})\right\|_{1}.
\end{align}

\subtitle{Scale-preserving Loss} Since we convert the sparse motions at the mesh vertexes as a dense motion field and predict the mesh warp field, we introduce a scale-preserving loss to maintain the scale consistency of the predicted warp field:
\begin{align}\label{eq:sp_loss}
   \mathcal{L}_{\text{SP}} = \frac{1}{N} \sum_{v_i}\left\| \frac{\left\| v_{i}^1 - v_{i} \right\|_{2}}{s} - 1 \right\|_{1}.
\end{align}
where $s$ is the scale factor in Eq.~\ref{eq:f_t}. Eventually, the overall loss is the combination of the above three loss terms:
\begin{align}\label{eq:loss}
   \mathcal{L} = \mathcal{L}_{\text{MC}} + \alpha \mathcal{L}_{\text{SC}} + \beta \mathcal{L}_{\text{SP}},
\end{align}
where $\alpha$ and $\beta$ are empirically set to 0.01.

\begin{figure*}[t]
   \centering
   \includegraphics[width=0.95\linewidth]{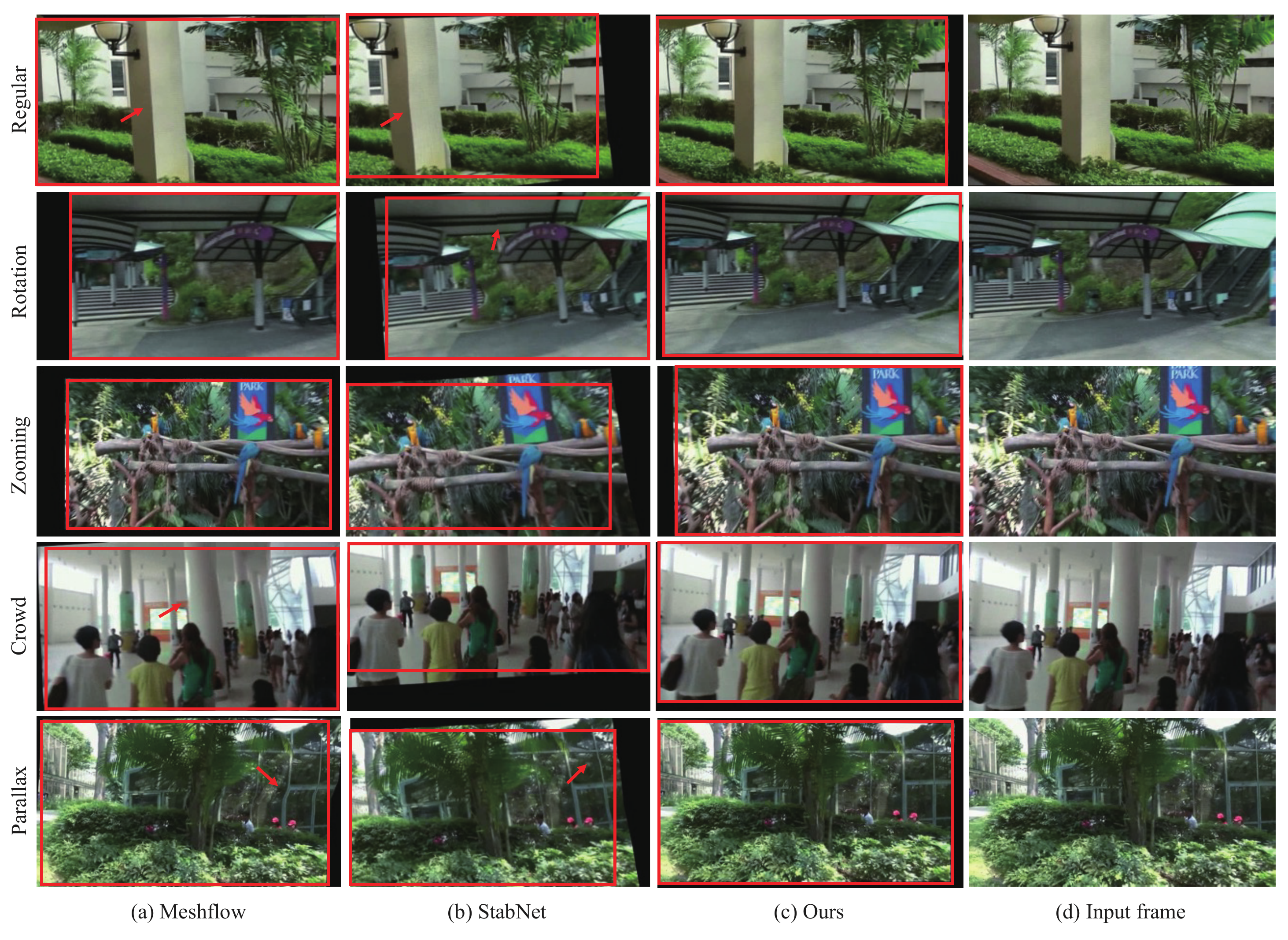}   
   \caption{Qualitative comparisons with the online methods Meshflow~\cite{liu2016meshflow} and StabNet~\cite{wang2018deep}. We keep the original warped frames for comparison. The red boxes indicate the preserved frame content after cropping. The red arrows highlight the shear and distortion artifacts.}\label{fig:result_nus}
 \vspace{-0.5em}
 \end{figure*}

\section{Experimental Results}

\subsection{Implementation Details}

\subtitle{Evaluation Metrics} For quantitative comparison, we employ three widely used metrics: \textbf{cropping ratio}, \textbf{distortion value}, and \textbf{stability score}, as in prior methods~\cite{liu2013bundled,xu2022dut,zhao2020pwstablenet,liu2016meshflow}. For cropping ratio and distortion value, we first fit a global homography between the input and output videos at each frame. The cropping ratio measures the remaining frame area after cropping out undefined pixels due to motion compensation and is computed as the average scale component of the entire video. The distortion value, which is defined as the worst ratio of the two largest eigenvalues of the affine component across all frames, measures the anisotropic scaling of the homography between the input and output frames. The stability score measures the smoothness of the stabilized videos and the camera path is utilized to determine the value. We compute this metric by using a frequency domain analysis as in Bundled~\cite{liu2013bundled}. 

\subtitle{Implementation Details} Our network is implemented by PyTorch. We use Adam as the optimizer for training, with an initial learning rate of $1\times 10^{-4}$ and no weight decay. We set $\beta_{1} = 0.9$, $\beta_{2} =0.999$, and $\epsilon = 1e-8$, respectively. We train the network for 100,000 iterations. The overall training time on two NVIDIA 1080Ti GPUs is 20 hours. During training, two adjacent frames and their corresponding historical frames are selected as a training sample. The MotionStab dataset comprises 65238 training samples in total. For inference, the overall framework can process a frame of size $640\times 360$ in real-time. Specifically, we spend 12 ms, 10 ms, and 4 ms to extract camera motion, smooth the camera path, and synthesize stable frames, respectively.

\subsection{Comparisons with Existing Methods}
To evaluate the proposed approach, we compare it with the state-of-the-art methods on the NUS dataset~\cite{liu2013bundled}. The compared methods include three traditional methods, Bundled~\cite{liu2013bundled}, Robust L1~\cite{grundmann2011auto}, Meshflow~\cite{liu2016meshflow}, and five deep learning based methods, DIFRINT~\cite{choi2020deep}, Yu~\emph{et al.}~\cite{yu2020learning}, PWStableNet~\cite{zhao2020pwstablenet}, DUT~\cite{xu2022dut}, and StabNet~\cite{wang2018deep}. It is worth noting that Meshflow~\cite{liu2016meshflow}, StabNet~\cite{wang2018deep}, and ours are online, while the rest are offline. The results of the compared methods are obtained from publicly accessible implementations with default parameters or pre-trained models.


\subtitle{Quantitative results} Table~\ref{tab:quantitative_results_nus} presents the quantitative results for each of the five categories (including regular, quick rotation, zooming, crowd, and parallax), as well as the average results across all categories. From Table~\ref{tab:quantitative_results_nus}, we can see that offline approaches generally outperform online methods, especially in challenging scenarios. In comparison to existing online approaches, our method achieves significant performance improvements. Taking Meshflow~\cite{liu2016meshflow} as the baseline, the cropping ratio and the distortion value of our method increase by 17.25\% and 12.18\%, respectively. Notably, our method can achieve approaching results to several compared offline methods.


\subtitle{Qualitative results} Fig.~\ref{fig:result_nus} illustrates the flaws that can be noticed directly from the video frames, including shear, distortion, and over-cropping. As can be seen,
Meshflow suffers from shear and distortion in some scenes due to its reliance on well-optimized parameter tweaking. The shear and over-cropping problems in StabNet~\cite{wang2018deep} are more severe because they directly employ video frames as input for mixed camera motion estimation and path smoothing, making the performance and generalization ability difficult to even approach the traditional approach Meshflow~\cite{liu2016meshflow}. On the contrary, since our network solely smooths and optimizes the camera path, regardless of image content, it is more generic on diverse challenging scenes and can avoid the aforementioned issues in an online setting.

\begin{figure*}[t]
   \centering
   \includegraphics[width=0.95\linewidth]{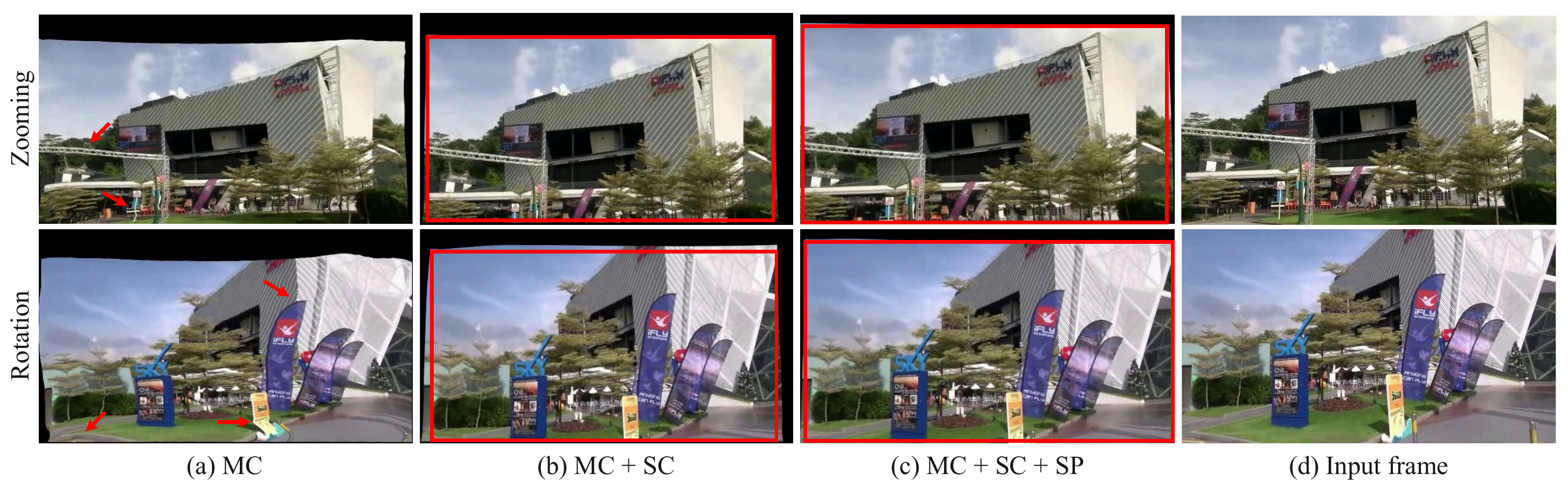}   
   \caption{Qualitative results of the ablation study on the motion-consistency loss (MC), the shape-consistency loss (SC), and the scale-preserving loss (SP). The red arrows show the shear and distortion regions, and the red boxes show the preserved frame region.}\label{fig:result_ablation_study_loss}
     \vspace{-1em}
 \end{figure*}

\begin{figure}[t]
  \centering
  \includegraphics[width=0.98\linewidth]{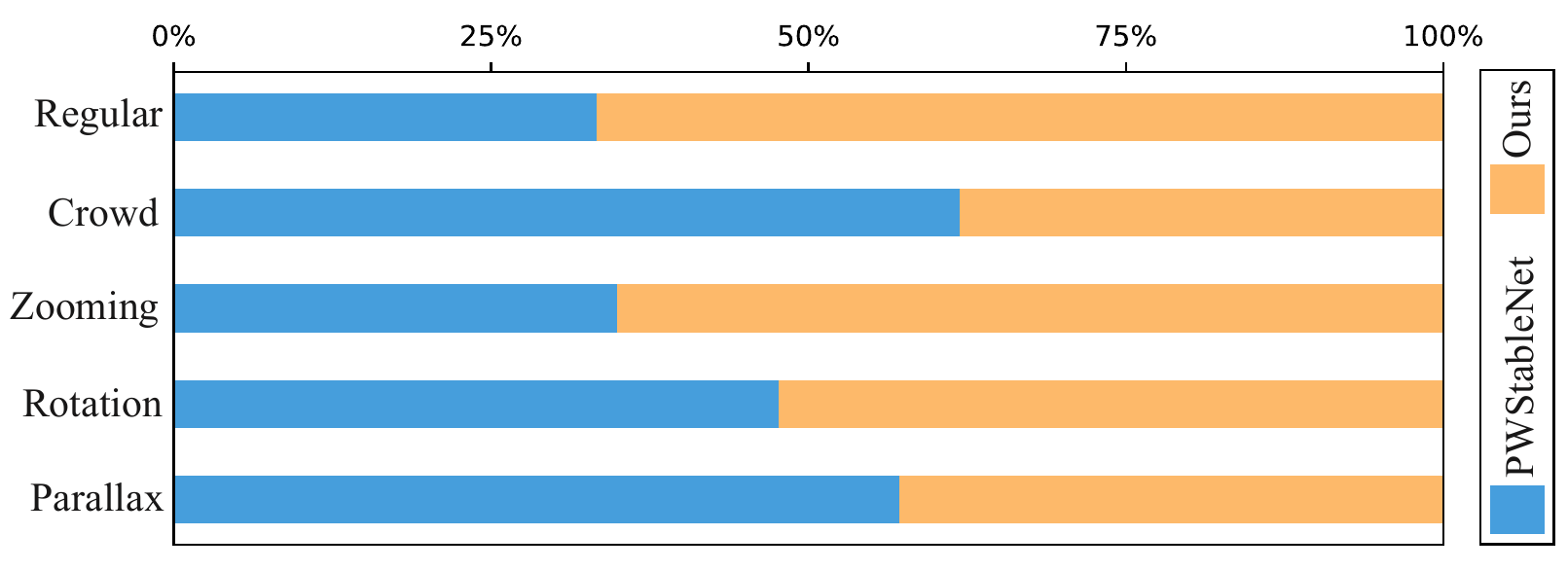}   
  \caption{User study results comparing with PWStableNet~\cite{zhao2020pwstablenet}.}\label{fig:user_study}
 \end{figure}

\subtitle{User Study} We also conduct a user study comparing our method to the PWStableNet~\cite{zhao2020pwstablenet}, which is one of the deep offline approaches that achieves overall good performances according to Table~\ref{tab:quantitative_results_nus}. Note that we do not compare with the online methods~\cite{liu2016meshflow,wang2018deep}, because our qualitative results are significantly better than theirs. Specifically, we prepare a test set containing 15 videos (3 videos per category), and each of the 20 participants is asked to choose the better one from the results of the two compared methods. The videos are arranged randomly and the original unstable videos are provided. The results of the user study are shown in Fig.~\ref{fig:user_study}.

\subsection{Ablation Studies}
To analyze the capability of each component, we conduct extensive ablation experiments on the NUS dataset~\cite{liu2013bundled}.

\subtitle{Ablation on losses} We first conduct experiments on the motion-consistency loss, the shape-consistency loss, and the scale-preserving loss. In the first row of Table~\ref{tab:quantitative_results_ablation}, we take the vanilla UNet optimized by the MC loss as our baseline. In the next two rows, we add the SC loss and the SP loss to train the model, respectively. Comparing the first three rows reveals that adding the SC loss and the SP loss respectively improves performance. The best performance is gained by combining them together. Fig.~\ref{fig:result_ablation_study_loss} shows some qualitative results. As seen, when only using the MC loss, substantial shear, and distortion occur (Fig.~\ref{fig:result_ablation_study_loss} (a), highlighted by red arrows.). The distortion artifacts can be effectively resolved by adding the SC loss (Fig.~\ref{fig:result_ablation_study_loss} (b)). However, the scale of the stabilized frame is not well preserved, resulting in larger cropped regions. This issue can eventually be solved by the proposed SP loss (Fig.~\ref{fig:result_ablation_study_loss} (c), distinguished by red boxes).

\begin{table}[t]
  \centering
  \caption{Quantitative results of the ablation studies on the hybrid loss and the channel attention mechanism (CA).}
  \label{tab:quantitative_results_ablation}
  \resizebox*{0.95\linewidth}{!}{
  \begin{tabular}{
      >{\arraybackslash}p{1.4cm} 
				   >{\centering\arraybackslash}p{0.8cm} 
            >{\centering\arraybackslash}p{0.8cm} 
            >{\centering\arraybackslash}p{0.8cm} 
            >{\centering\arraybackslash}p{0.8cm} 
				   >{\centering\arraybackslash}p{0.7cm} 
				>{\centering\arraybackslash}p{0.7cm} 
				>{\centering\arraybackslash}p{0.7cm} 
      }
  \toprule
             & MC & SC & SP & CA & C    & D    & S    \\ \midrule
  Baseline  & \checkmark   &    &    &    & 0.80 & 0.84 & 0.87 \\
  Variant 1 &  \checkmark   &  \checkmark   &    &    & 0.85 & 0.90 & 0.86 \\
  Variant 2 &  \checkmark   & \checkmark    &   \checkmark  &    & 0.90 & 0.95 & 0.86 \\
  Ours      & \checkmark    &   \checkmark  &  \checkmark   & \checkmark    & 0.92 & 0.98 & 0.87 \\ \bottomrule
  \end{tabular}}
  \vspace{-0.3em}
  \end{table}

\begin{figure}[t]
   \centering
   \includegraphics[width=0.95\linewidth]{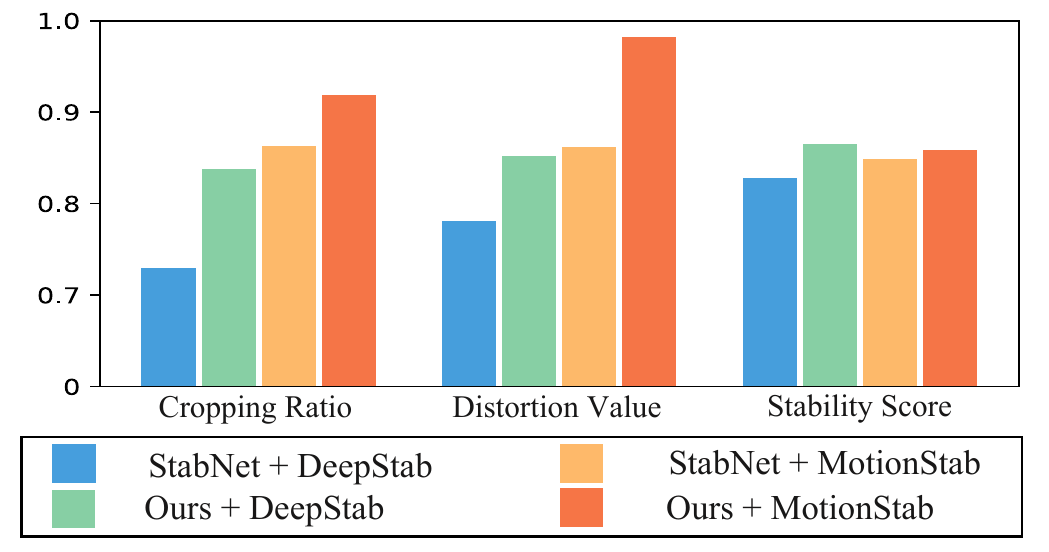}   
   \caption{Comparisons of the quantitative results between StabNet~\cite{wang2018deep} and our method when trained with different datasets.}\label{fig:dataset_analysis}
  \vspace{-1em}
 \end{figure}

\subtitle{Analysis of the quality of the dataset} To verify the impact of the dataset, we train the network on the DeepStab~\cite{wang2018deep} and our MotionStab dataset, respectively. Fig.~\ref{fig:dataset_analysis} shows the quantitative results on the NUS dataset. As seen, when trained using the proposed MotionStab dataset, both our method and StabNet~\cite{wang2018deep} yield better results than training on the DeepStab dataset, respectively, especially in terms of cropping ratio and distortion value. We conclude that the main reasons are twofold. Firstly, our MotionStab dataset contains richer scenarios and unstable motions, improving the performance and generalization ability of the network. Secondly, the synthesized pairs in MotionStab suffer no parallax issue inherent in DeepStab, leading to fewer distortion artifacts (i.e., higher distortion value) in the stabilized frames. Additionally, our method outperforms StabNet~\cite{wang2018deep} on both datasets, demonstrating its superiority over directly predicting stable warp fields from input video frames.

\section{Conclusions}
This work presents a deep camera path optimization framework for online video stabilization. We leave the motion estimation to recent off-the-shelf deep motion models and concentrate on path smoothing. Our model takes a 2D camera path in a sliding window as input and outputs a warp field for the last frame in that window. A video can be stabilized online by applying our model repeatedly for every incoming frame. We introduce a hybrid loss to enhance the spatial and temporal consistency of the stabilized video. Moreover, we created a motion dataset to train our model. Results show that our method outperforms previous approaches both qualitatively and quantitatively. 

\noindent\textbf{Acknowledgements} This work was supported by Sichuan Science and Technology Program of China under grants Nos.2023NSFSC0462, 2023NSFSC1972, 2022YFQ0079, 2021YFG0001, and National Natural Science Foundation of China (NSFC) under grant No.62031009.

{\small
\bibliographystyle{ieee_fullname}
\bibliography{nndvs}
}

\end{document}